\documentclass[11pt,a4paper]{article}
\usepackage[hyperref]{acl2021}
\usepackage{times}
\usepackage{float}
\usepackage{latexsym}
\usepackage{url}
\usepackage{multirow}
\usepackage[ruled,vlined]{algorithm2e}
\usepackage{algorithmic}
\usepackage[utf8]{inputenc} 
\usepackage[T1]{fontenc} 
\usepackage{hyperref} 
\usepackage{url} 
\usepackage{booktabs} 
\usepackage{amsfonts} 
\usepackage{nicefrac} 
\usepackage{microtype} 
\usepackage[utf8]{inputenc} 
\usepackage[T1]{fontenc} 
\usepackage{hyperref} 
\usepackage{url} 
\usepackage{booktabs} 
\usepackage{amsfonts} 
\usepackage{nicefrac} 
\usepackage{microtype} 
\usepackage{graphicx}
\usepackage{xcolor}
\usepackage{gensymb}
\usepackage{subcaption}
\usepackage{amsthm}
\usepackage{mathrsfs}
\usepackage{amsmath}
\usepackage{amssymb}
\usepackage{mathtools}
\usepackage{wrapfig}
\usepackage{soul}
\makeatletter
\def\BState{\State\hskip-\ALG@thistlm}
\makeatother

\newcommand{\ba}[1]{\begin{align}#1\end{align}}
\newcommand{\baa}[1]{\begin{equation}\begin{aligned}#1\end{aligned}\end{equation}}

\usepackage{stackengine}

\makeatletter
\newcommand{\distas}[1]{\mathbin{\overset{#1}{\kern\z@\sim}}}%

\newcommand{\cD}{\mathcal{D}}

\newcommand{\cM}{\mathcal{M}}

\usepackage{algorithm2e}

\newcommand{\bx}{\boldsymbol{x}}

\newcommand{\beqs}{\vspace{0mm}\begin{eqnarray}}
\newcommand{\eeqs}{\vspace{0mm}\end{eqnarray}}
\newcommand{\barr}{\begin{array}}
\newcommand{\earr}{\end{array}}

\newcommand{\xv}{\boldsymbol{x}}

\usepackage{bbm}

\newcommand{\ours}{{ALLSH}}
\newcommand{\med}[1]{\textcolor{magenta}{#1}}
\newcommand{\aux}[1]{~~~~~~~\text{\med{//#1}}}

\title{
ALLSH: Active Learning Guided by Local Sensitivity and Hardness
}

\author{Shujian Zhang$^{1,2}$, Chengyue Gong$^1$, Xingchao Liu$^1$, Pengcheng He$^2$,
\\
\bf{Weizhu Chen$^2$, Mingyuan Zhou$^{1}$}\\
$^1$The University of Texas at Austin \quad \quad $^2$Microsoft Azure AI
\\
\texttt{\{szhang19, cygong, xcliu\}@utexas.edu}
\\\texttt{\{penhe, wzchen\}@microsoft.com}
\\\texttt{mingyuan.zhou@mccombs.utexas.edu}
}

\begin{document}
\maketitle

\begin{abstract} 
Active learning, which effectively collects informative unlabeled data for annotation, reduces the demand for labeled data. In this work, we propose to retrieve unlabeled samples with a local sensitivity and hardness-aware acquisition function. The proposed method generates data copies through local perturbations and selects data points whose predictive likelihoods diverge the most from their copies. We further empower our acquisition function by injecting the select-worst case perturbation. Our method achieves consistent gains over the commonly used active learning strategies in various classification tasks. Furthermore, we observe consistent improvements over the baselines on the study of prompt selection in prompt-based few-shot learning. These experiments demonstrate that our acquisition guided by local sensitivity and hardness can be effective and beneficial for many NLP tasks.\let\thefootnote\relax\footnote{Code is available at \url{https://github.com/szhang42/allsh}} 


\end{abstract}

\section{Introduction}\label{sec:introduction}
Crowdsourcing annotations~\cite{Rajpurkar2016SQuAD10,Bowman2015ALA} has become a common practice for developing NLP benchmark datasets. Rich prior works~\cite{Pavlick2019InherentDI,nie2020can,ferracane-etal-2021-answer} show that the time-consuming and expensive manual labeling in  crowdsourcing annotations are not an annotation artifact but rather core linguistic phenomena. Active Learning (AL) is introduced to efficiently acquire data for annotation from a (typically large) pool of unlabeled data. 
Its goal is to concentrate the human
labeling effort on the most informative data
in hopes of maximizing the  model performance  while minimizing the data annotation cost.

Popular approaches to acquiring data for AL are uncertainty sampling and diversity sampling. Uncertainty sampling selects data
that the model predicts with low-confidence \cite{lewis1994sequential, culotta2005reducing, settles2009active}. Diversity sampling selects
batches of unlabeled examples that are prototypical of the unlabeled pool to exploit heterogeneity in the feature space \cite{xu2003representative, bodo2011active}. 
Different from these two perspectives, recent works focus on the informativeness of the selected data.
For example, \citet{zhang2021cartography} acquire informative unlabeled data using the training dynamics based on the model predictive log likelihood.  \citet{margatina2021active} construct contrastive examples in the input feature space. However, these methods either ignore the local sensitivity of the input features or take no consideration of the difficulty of the learning data. 
Consequently, they may ignore examples around the decision boundary, or select hard-to-train or even noisy examples.
Their performance may further suffer under some practical settings, such as those with imbalanced labels and when there is a very limited annotation budget. 

\begin{figure}[t]
\centering
\includegraphics[width=7.5cm]{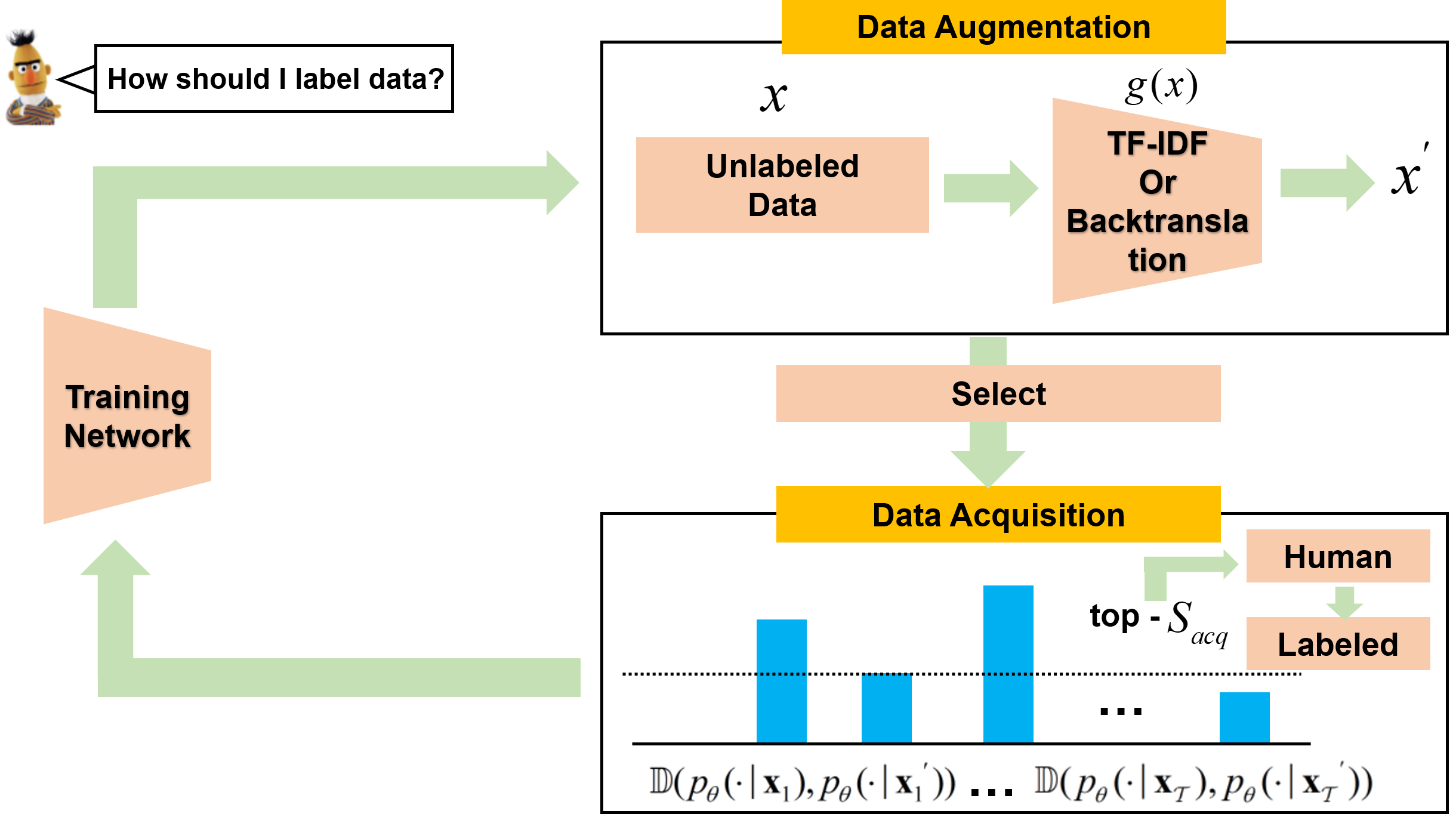}
\vspace{-3mm}
\caption{Overview of active learning framework guided by local sensitivity and hardness. Some notations are labeled along with corresponding components. `Select' refers to the select worst-case augmentation. 
}
 \vspace{-3mm}
\label{fig:pipeline}
\end{figure}

In this work, we determine the informativeness by considering both the local sensitivity and learning difficulty. For local sensitivity, we take the classical definition from \citet{chapelle2009semi}, which is widely used in both classic machine learning problems \citep[e.g.][]{blum2001learning,chapelle2002cluster,seeger2000learning,zhu2003semi, zhou2004learning} and recent deep learning settings \citep[e.g.][]{wang2018identifying, sohn2020fixmatch, xu2021locality}. 
Specifying a local region $\mathcal{R}_\textit{region}(\bx)$ around an example $\bx$, we assume in our prior that all examples in $\mathcal{R}_\textit{region}(\bx)$ have the same labels.\footnote{See the paragraph `unlabeled bias as regions' and the section `Regions and Smoothness' for details.}
If the examples in $\mathcal{R}_\textit{region}(\bx)$ give us different labels, we say the local region of $\bx$ is sensitive.
Data augmentation has been chosen as the way to create label-equivalent local regions
in many recent works \citep[$e.g.$,][]{berthelot2019mixmatch, Xie2020UnsupervisedDA}. 
We utilize data augmentation as a tool to capture the local sensitivity and hardness of inputs and 
present {\ours}: Active Learning guided by Local Sensitivity and Hardness.
Through various designs on local perturbations, {\ours} selects unlabeled data points from the pool whose predictive likelihoods diverge the most from their augmented copies. This way, {\ours} can effectively ensure the informative and local-sensitive data to have correct human-annotated labels. Figure \ref{fig:pipeline} 
illustrates the scheme of the proposed acquisition strategy.

We conduct a comprehensive evaluation of our approach on datasets ranging from sentiment analysis, topic classification, natural language inference, to paraphrase detection. 
To measure the proposed acquisition function in more realistic settings where the samples stem from a dissimilar input distribution, we (1) set up an out-of-domain test dataset and (2) leak out-of-domain data ($e.g.$, adversarial perturbations) into the selection pool. 

We further expand the proposed acquisition to a more challenging setting: prompt-based few-shot learning~\cite{zhao2021calibrate}, where we query a fixed pre-trained language model via a natural language prompt containing a few training examples. 
We focus on selecting the most valuable prompts for a given test task ($e.g.$, selecting 4 prompts for one given dataset).
We adapt our acquisition function  to retrieve prompts for the GPT-2 model. 

Furthermore, we provide extensive ablation studies on different design choices for the acquisition function, including the designs of augmentations and divergences.
Our method shows consistent gains in all settings with multiple datasets. With little modification,
our data acquisition can be easily applied to other NLP tasks for a better sample selection strategy. 

Our contributions are summarized as follows: (1) Present a new acquisition strategy, embracing local sensitivity and learning difficulty, such as paraphrasing the inputs through data augmentation and adversarial perturbations, into the selection procedure. (2) Verify the effectiveness and general applicability of the proposed method in more practical settings with imbalanced datasets and extremely few labeled data. (3) Provide comprehensive study and experiments of the proposed selection criteria in classification tasks (both in-domain and out-of-domain evaluations) and prompt-based few-shot learning. (4) The proposed data sampling strategy can be easily incorporated or extended to many other NLP tasks.

\section{Method}
In this section we present in detail our proposed
method, {\ours} (Algorithm \ref{alg:acquisition}).

\subsection{Active Learning Loop}
The active learning setup consists of an unlabeled dataset $\cD_{pool}$, the current training set $\cD_{label}$, and a model $\mathcal M$ whose output probability is $p_\theta(\cdot \mid \bx)$ for input $\bx$. The  model $\cM$ is generally a 
pre-trained model for NLP tasks \cite{lowell2018practical}. 
At each iteration, we train a model on $\cD_{label}$
and then use the acquisition function to acquire $s_\textit{acq}$ sentences in a batch $\mathcal{T}$ from $\cD_{pool}$. The acquired examples from this iteration are labeled, added to $\cD_{label}$, and removed from $\cD_{pool}$. Then the updated $\cD_{label}$ serves as the training set in the next AL iteration until we exhaust
the budget. 
Overall, the system is given a budget of $S$ queries to build a labeled training dataset of size $S$. 

\subsection{Acquisition Function Design} \label{sec:acq_design}

To fully capture the data informativeness and train a model with a limited amount of data, 
we consider two data-selection principals: local sensitivity and learning hardness. 

\noindent {\bf Local Sensitivity} 
Based on theoretical works on the margin theory for active learning, the examples lying close to the decision boundary are informative and worth labeling \cite{ducoffe2018adversarial,margatina2021active}. 
Uncertainty sampling suffers from the sampling bias problem as the model is only trained with few examples in the early phase of
training. In addition, high uncertainty samples given
the current model state may not  be that representative
to the whole unlabeled data \cite{ru2020active}.
For example, 
if an input has high confidence while its local perturbation generates low-confidence output, then it is likely that this input lies close to the model decision boundary.
This information can be captured by measuring the difference between an input example and its augmentation in the output feature space.
We utilize the back-translation \cite{Sennrich2016ImprovingNM, edunov2018understanding,zhang2021knowing} and TF-IDF \cite{Xie2020UnsupervisedDA} as effective augmentation methods which can generate diverse paraphrases while preserving the semantics of the original inputs \cite{yu2018qanet}. 

Instead of simply using augmentation, adversarial perturbation can measure the local Lipschitz and sensitivity more effectively.
We therefore further exploit adversarial perturbation to more accurately measure local sensitivity.
For NLP problems, generating exact adversarial perturbations in a discrete space usually requires combinatorial optimization, which often suffers from the curse of dimensionality 
~\citep{madry2017towards,lei2018discrete}.
Hence, we choose the hardest augmentation over $K$ random augmentations as a “lightweight” variant of adversarial input augmentation which optimizes the worst case loss over the augmented~data. 

\noindent  {\bf Learning Hardness: From Easy to Hard } 
Learning from easy examples or propagating labels from high-confidence examples is the key principle for curriculum learning \citep{bengio2009curriculum} and label propagation based semi-supervised learning algorithms \citep{chapelle2009semi}.
For example,  FixMatch \citep{sohn2020fixmatch}, a SOTA semi-supervised method, applies an indicator function to select high confident examples at each iteration. 
This will facilitate the label information from high confidence examples to low-confidence ones \cite{chapelle2009semi}.
In our selection criterion,
as the model is trained with limited data, 
we also want to avoid the hard-to-learn examples, which 
in some cases 
frequently correspond to mislabeled
or erroneous instances \cite{Swayamdipta2020DatasetCM,zhang2021cartography}. 
These examples may stuck the model performance at the beginning of the selection.

\subsection{Acquisition with Local Sensitivity and Hardness}\label{sec:acquisition_with_local}

We come to the definition of our acquisition function.
Given a model $p_\theta$ and an input $\bx$, we compute the output distribution $p_\theta( \cdot\mid \bx)$ and a noised version $p_\theta( \cdot \mid \bx^{\prime})$ by injecting a random transformation $\bx^{\prime} =g(\bx)$ to the inputs. 
Here, $g(\cdot)$ is sampled from a family of transformations and these random transformations stand for  data augmentations. This procedure can select examples that are insensitive to transformation $g(\cdot)$ and hence smoother with respect to
the changes in the input space \cite{berthelot2019mixmatch, berthelot2019remixmatch, sohn2020fixmatch}. We calculate
\vspace{-1mm}
\baa{
\ell(\bx, \bx') = \mathbb{D}(p_\theta(\cdot \mid \boldsymbol{x}), p_\theta(\cdot \mid \boldsymbol{x^{\prime}})),
\vspace{-3mm}
\label{eq:acqui}
}
where $\mathbb{D}$ denotes a statistical distance such as the Kullback–Leibler (KL) divergence \cite{kullback1951information}. Model $p_\theta$ here can be a pretrained language model such as BERT \cite{devlin2018bert}.  


\paragraph{Data Paraphrasing via Augmentation} Paraphrase generation can improve language models~\cite{Yu2018QANetCL} by handling language variation. TF-IDF and backtranslation can generate diverse  inputs while preserving the semantic meaning \cite{singh2019xlda, Xie2020UnsupervisedDA}. For TF-IDF, we replace uninformative words with low TF-IDF scores while keeping those with high. Specifically, Suppose IDF$(w)$ is the IDF score for word $w$ computed on the whole corpus, and
TF$(w)$ is the TF score for word $w$ in a sentence. We compute the TF-IDF score as TFIDF$(w) =$
TF$(w)$IDF$(w)$. For backtranslation, we use a pre-trained EN-DE and DE-EN translation
models \cite{ng2019facebook} to perform backtranslation on each sentence. We denote $\boldsymbol{x}$ as $(x_0, \cdots, x_{n})$.
Here, $n$ denotes the original length of the input.
For $\boldsymbol{x}$, we pass 
them through two translation models to get $\bx' = (x'_0, \cdots, x'_{m})$, where $m$ denotes the length after backtranslating.
More details can be found in Appendix \ref{sec:app_exp}.

\paragraph{Select Worst-Case Augmentation (WCA)}

In order to construct effective local sensitivity, the most direct approach is calculating the local Lipschitz constant or finding the worst case adversarial perturbation.
However, estimating the Lipschitz constant for a neural network is either model dependent or computationally
hard \citep{scaman2018lipschitz, fazlyab2019efficient}. 
Instead, we select worst-case augmentation over $K$ copies, which can still roughly measure the norm of the first-order gradient without a huge computation cost and is easy to implement.
Given input examples $\bx$, and $K$ augmentation of $\bx$ as $\{\bx_i'\}_{i=1}^{K}$, 
we propose the following acquisition function to select data:

\vspace{-3mm}
\begingroup\makeatletter\def\f@size{10}\check@mathfonts
\def\maketag@@@#1{\hbox{\m@th\large\normalfont#1}}%
\baa{
\ell^\textit{max}(\bx) =  \max_{i\in[K]} \ell(\bx, \bx_i').
\vspace{-3mm}
\label{eq:acqui-max}
}
\endgroup
%
Inspired by some simple and informal analysis in continuous space, 
we draw the connection between calculating $\ell^\textit{max}(\bx)$ and local sensitivity by

\vspace{-6mm}
\begin{align}
\begin{split}
    & \resizebox{0.76\hsize}{!}{$\ell^\textit{max}(\bx) = \ell(\bx, \bx') + \bigg [ \ell^\textit{max}(\bx) - \ell (\bx, \bx') \bigg]$} \\
    & \resizebox{0.9\hsize}{!}{$= \ell (\bx, \bx') + 
     \left[ \max_{i\in[K]}\langle \nabla_{\bx} \ell (\bx, \bx'), \bx - \bx' \rangle \right]+ \mathbf O(\sigma^2).$}
\end{split}\!
\end{align}
Recent works in computer vision \citep{gong2020maxup, wang2021augmax} have provided  more formal 
connections between local gradient norm estimation and $K$-worst perturbations. 

The text sentences in NLP are in the discrete space, which lacks the definition of local Lipschitz, but  finding the worst perturbation in a local discrete set can still be a better measurement of local sensitivity in the semantic space. 

\noindent {\bf Choice of Divergence}
We use the KL divergence 
as the primary measure of the statistical distance between the distribution of the original examples and
that over augmented examples. We also empirically provide detailed analysis of the Jensen--Shannon Distance (JSD) \cite{endres2003new} and $\alpha$-divergence \cite{minka2005divergence} as a complementary measure in Section \ref{sec:analysis_section}. The
$\alpha$-divergence \cite{pillutla2021mauve} is a general divergence family, which includes the most popular KL divergence and reverse KL divergence. 
Different value of $\alpha$ makes the divergence trade-off between  overestimation and underestimation. 
JSD is
a metric function based on a mathematical definition which is symmetric
and bounded within the range $[0, 1]$.
These divergences are calculated as:
\begin{align}
    \begin{split}
        & \resizebox{0.5\hsize}{!}{${\mathrm{KL}} (p\|q) =
       \sum\limits_{i} p_i(\bx) \log \frac{p_i(\bx)}{q_i(\bx)},
        $} \\
        & \resizebox{0.6\hsize}{!}{${\mathrm{JSD}} (p\|q) = \sqrt{\frac{1}{2}(\mathrm{KL}(p \| m)+\mathrm{KL}(q \| m))},$} \\
        & \resizebox{0.6\hsize}{!}{$\mathrm{D}_{\alpha} (p\|q) = \frac{1}{\alpha(\alpha-1)} \sum\limits_{i}[(\frac{p_i(\bx)}{q_i(\bx)})^{\alpha}-1],$} \\
    \end{split}
\end{align}
where $p$ is the output probability distribution of an example, $q$ is the output probability
distribution of an augmented example, and $m=\frac{1}{2}(p+q)$.

\noindent {\bf Local Sensitivity and Informativeness}
\begin{figure}[t]
\centering
\includegraphics[width=3.8cm]{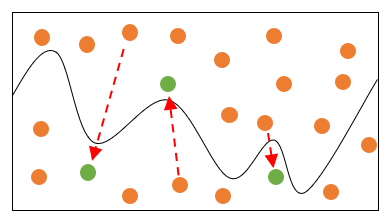}
\vspace{-3mm}
\caption{The solid line is model decision boundary. Orange circles refer to the unlabeled data and green circles refer to the corresponding augmentation of the orange unlabeled data. 
}
 \vspace{-10mm}
\label{fig:local_sensitivity}
\end{figure}
The divergence objective exploits unlabeled data by measuring
predictions across slightly-distorted versions of each unlabeled sample. The diverse and adversarial augmentations capture the local sensitivity and informativeness of inputs and project examples to the decision boundary  \cite{ducoffe2018adversarial}. Thus,  the examples and their copies with highly inconsistent model predictions lie close to the decision boundary of the model \cite{gao2020consistency}. These examples are valuable to have human annotations because they 1) contain high-confidence region in a local perturbation and are therefore easy to train 2) are highly likely to promote the model with large-margin improvements (see example in Figure~\ref{fig:local_sensitivity}). Under our local sensitivity and hardness guided acquisition, we argue
the selected examples would not be necessarily the examples with the highest uncertainty, which do not always benefit the training.
For instance, an example may have low-confidence prediction of both original inputs and augmented inputs thus making the samples most hard to train.

\subsection{More Details}
{\bf Compute Distance}
We compute the divergence in the model predictive probabilities for the pairs of the input and its augmentations in Eqn \eqref{eq:acqui}. Specifically, we use a pretrained BERT in classification tasks and GPT-2 in prompt-based few-shot learning as the base model $p_\theta$ to
obtain the output probabilities for all unlabeled data points in 
$\cD_{pool}$. We then compute the divergence value with Eqn \eqref{eq:acqui}.
{\bf Rank and Select Candidates} We apply these steps to all candidate examples from $\cD_{pool}$ and obtain the divergence value for each. 
Our acquisition function selects the top $s_\textit{acq}$ examples
that have the highest divergence value from the acquired batch~$\mathcal{T}$.

\begin{algorithm*}[t]
\caption{\small Acquisition with Local Sensitivity and Hardness} 
\label{alg:acquisition}
\begin{algorithmic}[1]
\footnotesize 
\STATE \textbf{Input:} labeled data $\cD_{label}$, unlabeled data $\cD_{pool}$, acquisition size $s_\textit{acq}$, model $\mathcal M$ with output probability $p_\theta(\cdot \mid \bx)$.
\STATE \textbf{while} ~Select examples before reaching the budget~ \textbf{do}
\FOR{$\bx$ in  $\cD_{pool}$}
\STATE Generate $K$ augmentations,  $\left\{\bx_{i}^{\prime}\right\}_{i=1, \cdots, K} \leftarrow g\left(\bx\right)$. \aux{data paraphrasing via augmentation}
\STATE Compute $p_\theta(\cdot \mid \bx)$ and  $p_\theta(\cdot \mid \bx^{\prime}_{i})$ for $i=1, \ldots, K$. \aux{compute probabilities}
\STATE
Select the worst case augmentation $\xv'$ for each input $\bx$ as $\ell^\textit{max}(\bx) =  \max_{i=1, \cdots, K} \ell(\bx, \bx_i')$.
 \ENDFOR
 \STATE
Select top $s_\textit{acq}$ largest examples in $\cD_{pool}$, according to the value of $\mathbb{D}(p_\theta(\cdot \mid \boldsymbol{x}), p_\theta(\cdot \mid \boldsymbol{x^{\prime}}))$.
 \STATE Label these $s_\textit{acq}$ examples.
 \STATE \textbf{end while}
\STATE Curriculum learning the model parameters with  
Eqn \eqref{eq:ssl}.
\end{algorithmic}
\end{algorithm*}

\section{Experimental Settings}\label{sec:experiemental_settings}

Table \ref{tab:dataset_setting} shows the  experimental data configuration. In classification tasks, we use five datasets, including Stanford Sentiment Treebank (SST-2; \cite{socher2013recursive}), Internet Movie Database (IMDB; \cite{maas2011learning}), AG’s News
Corpus (AG News; \cite{zhang2015character}),
Quora Question Pairs (QQP; \cite{wang2018glue}), and Question NLI (QNLI; \cite{wang2018glue}). 
The validation and test splits are provided in \citet{ margatina2021active}. 
Following~\citet{Desai2020CalibrationOP}, we test domain generalization and robustness on three challenging out-of-domain (OD) datasets.
For sentiment analysis, 
SST-2 and IMDB  are the source and target domains, respectively, and vice versa; 
for paraphrase detection, TwitterPPDB \cite{lan2019albert} serves as the out-of-domain test dataset for~QQP.


\begin{table}[t]
\centering
\footnotesize
\resizebox{0.8\columnwidth}{!}{
 \begin{tabular}{l|c|c|c|c}
 \toprule
 \bf{Dataset} &   \bf{Train} & \bf{Val}  & \bf{Test} & \bf{OD Dataset} \\ \midrule
  SST-2 & 60.6K & 6.7K & 871 & IMDB \\
 IMDB & 22.5K & 2.5K & 25K & SST-2  \\
 AG News & 11.4K & 6K & 7.6K & -  \\
 QNLI & 99.5K & 5.2K & 5.5K & - \\
 QQP & 327K & 36.4K & 80.8K & TwitterPPDB  \\
\hline \hline \\ [-2.0ex]
 SST-2  & 60.6K & 6.7K & 871 & - \\
 TREC  & 4.5K & 500 & 500 & - \\
 RTE  & 2.5K & 277 & 3K & - \\
 \bottomrule
  \end{tabular}}
 \caption{Dataset Configuration. The top block is for the classification tasks and the bottom block is for the prompt-based few-shot learning. OD represents out-of-domain datasets.} 
   \vspace{-4mm}  
    \label{tab:dataset_setting}
\end{table}

In the prompt-based few-shot learning, we follow \citet{zhao2021calibrate} to use SST-2 \cite{socher2013recursive} for sentiment analysis, TREC \cite{voorhees2000building} for  question classification,  and RTE 
\cite{dagan2005pascal} for recognizing textual entailment. 
See Appendix~\ref{sec:app_exp} for more details of the~data. 


\subsection{Classification Task}
We compare the proposed {\ours~} against four baseline methods. We choose these baselines as they cover a spectrum of acquisition functions (uncertainty, batch-mode, and diversity-based).

\noindent {\bf Random } samples data from the pool of unlabeled data $\cD_{pool}$ following a uniform distribution.

\noindent {\bf Entropy } selects $s_\textit{acq}$  sentences with the highest predictive entropy \cite{lewis1994sequential} measured by
$
 - \sum\limits_{\bx}^{} p_{\theta}(\bx) \ln p_{\theta}(\bx).
$

\noindent {\bf BADGE \cite{Ash2020Deep}} acquires $s_\textit{acq}$  sentences based on diversity
in loss gradient. The goal of BADGE is to sample a diverse and
uncertain batch of points for training neural networks. It acquires data from $\cD_{pool}$ by first passing the input through the trained model and computing the gradient embedding with respect to the parameters of the model’s last layer.

\noindent {\bf CAL \cite{margatina2021active}} The acquisition function samples contrastive examples. It uses information from
the feature space to create neighborhoods for
unlabeled examples, and uses predictive likelihood for
ranking the candidates.

\subsection{Prompt-based Few-Shot Learning}
Following \citet{zhao2021calibrate}, we adapt our acquisition function for state-of-the-art generation based model GPT-2 and propose to retrieve examples that are semantics and sensitivity aware to formulate its corresponding prompts. We compare {\ours}'s acquisition function with \textit{random, contextual calibrated, and uncertainty prompt}. For random prompt, we
randomly select in-context examples from the
training set for each test sentence. 
For \textit{Calibrated}, \citet{zhao2021calibrate} inject calibration parameters that cause the prediction for each test input to be uniform across answers. See \citet{zhao2021calibrate} for more details.
For \textit{Uncertain}, we sample the highest uncertain prompt for the test sentences. For \textit{\ours}, we augment the in-context examples and select the prompts with the highest divergence of the predicted likelihood between the original examples and their augmentations.

\subsection{Implementation Details}\label{sec:implementation_details}
For  classification, we use BERT-base \cite{devlin2018bert} from the HuggingFace library \cite{wolf2020transformers}. We
train all models with batch size $16$, learning rate
$2 \times 10^{-5}$, and AdamW optimizer with epsilon $1 \times 10^{-8}$. For all datasets, we set the default annotation budget as $1\%$, the maximum annotation budget as $15\%$, initial accumulated labeled data set $\cD_{label}$ as $0.1\%$ of the
whole unlabeled data, and acquisition size as $50$ instances for each active learning iterations, following prior work \citep[$e.g.$,][]{gissin2019discriminative,dor2020active,ru2020active}. 
{\bf Curriculum Learning (CL)} We further 
combine our acquisition function with advances in semi-supervised learning (SSL) \cite{berthelot2019remixmatch,sohn2020fixmatch}, which also integrates abundant unlabeled data into learning.
A recent line of work in SSL utilizes data augmentations, such as TF-IDF and  back-translation, to enforce local consistency of the model \citep{sajjadi2016regularization, miyato2018virtual}.
Here SSL can further distill information from unlabeled data and gradually propagate label information from labeled examples to unlabeled one during the training stage \cite{Xie2020UnsupervisedDA, zhang2021learning}. We construct the overall loss function as 
%
\ba{
\resizebox{0.8\hsize}{!}{$
\mathcal{L}= \mathcal{L}_{S} + \alpha \cdot \underbrace{ \mathbb{E}_{\bx \sim \mathcal{D}_{pool}}\mathbb{D}(p_\theta(\cdot \mid \boldsymbol{x}), p_\theta(\cdot \mid \bx'))}_{\mathcal{L}_{U}}$},
\label{eq:ssl}
}
%
%
where $\mathcal{L}_{S}$ is the cross-entropy supervised learning loss over labeled samples, $\mathcal{L}_{U}$
is the consistency regularization term, and $\alpha$ is a coefficient  \cite{tarvainen2017mean, berthelot2019mixmatch}.

For prompt-based few-shot learning,  we run experiments on 1.5B-parameters GPT-2 \cite{radford2019language}, a Transformer \cite{vaswani2017attention}  based language model. 
It largely follows the details of the OpenAI GPT model \cite{radford2018improving}.
We take the TF-IDF as the default augmentation method and provide a rich analysis of other augmentation methods in Section \ref{sec:analysis_section}. More detailed experimental settings are included in Appendix \ref{sec:app_exp}.



\section{Experiments}\label{sec:experiemental_results}
We evaluate the performance of our acquisition
and learning framework in this section. We bold the best results within Random, Entropy, BADGE, CAL, and the proposed ALLSH (Ours)
in tables. Then, we bold the best result within each column block. All experimental results are obtained with five independent runs
to determine the variance. See Appendix~\ref{sec:app_exp} for the full results with error bars.

\subsection{In-Domain Classification Task Results}\label{sec:in_domain_section}
In Table~\ref{tab:in_domain}, we evaluate the impact of our acquisition function under three different annotation budgets ($1\%$, $5\%$, and $10\%$). With a constrained  annotation budget, we see substantial gains on test accuracy with our proposed acquisition: {\ours} and selecting worst-case augmentation.
With this encouraging initial results, we further explore our acquisition with curriculum learning.  Across all settings, {\ours} is consistently the top performing method especially in $\text{SST-2}$, IMDB, and AG News. With a tight budget, our proposed acquisition can successfully integrate the local sensitivity and learning difficulty to generate annotated data. 

For BADGE, despite combining both uncertainty and diversity sampling, it only achieves the comparable results on QNLI, showing that gradient computing may not directly benefit  data acquisitions. 
In addition, requiring clustering for high dimensional data, BADGE is computationally heavy as its complexity grows exponentially with the acquisition size \cite{yuan2020cold}. We provide rich analysis of the sampling efficiency and running time for each method in Appendix~\ref{sec:app_exp} and include the results in  Table \ref{tab:appendix_runningtime}.
Also, \ours~ outperforms the common uncertainty sampling in most cases. 
Given the current model state,  uncertainty sampling chooses the samples 
that are not representative
to the whole unlabeled data, leading to ineffective sampling. CAL has an effective contrastive acquiring on QNLI.  We hypothesize that due to the presence of lexical and syntactic ambiguity between a pair of sentence, the contrastive examples can be used to push away the inputs in the feature space.
\begin{table}[h]
 \footnotesize
\centering
\resizebox{1.\columnwidth}{!}{
 \begin{tabular}{l|l|c|c|c}
 \toprule
  &Acquired dataset size:& \multicolumn{1}{c|}{1\%} &\multicolumn{1}{c|}{5\%} &\multicolumn{1}{c}{10\%}\\ 
   \hline
 \multirow{6}{*}{SST-2} & Random & 84.11& 86.53 & 88.05\\
 & Entropy & 84.53 & 87.82 & 89.45 \\
  & BADGE & 84.32& 87.11 & 88.72 \\
 & CAL & 84.95 & 87.34 & 89.16 \\
\cline{2-5}
 & {\bf Ours} & {\bf 85.97} & {\bf  88.61} & {\bf 90.05}\\
& + {\bf WCA} & 86.12 & 88.56 & 90.14\\
 & + {\bf CL} & {\bf 86.37} & {\bf 88.79}& {\bf 90.18}\\ 
  
 \hline
 \hline
 \multirow{6}{*}{IMDB}& Random & 65.90 & 84.22 & 86.25\\
 & Entropy & 68.32 & 84.51& 87.29 \\
 & BADGE & 67.80 & 84.46 & 87.17 \\
 & CAL & 73.55 & 84.72 & 87.27 \\
 \cline{2-5}
 & {\bf Ours:} & {\bf 75.23} & {\bf 85.82} & {\bf 87.91} \\
& + {\bf WCA} & 75.17 & 85.79 & 87.83\\
 & +  {\bf CL} & {\bf 77.57} & {\bf 86.02}  & {\bf 88.43} \\ 
 \hline
 \hline
 \multirow{6}{*}{AG News} &Random & 85.43 & 90.05 & 91.93\\
 &Entropy & 86.48 & 92.21 & 92.65 \\
 &BADGE & 86.81  & 90.72 & 92.41 \\
 & CAL & 87.12 & 92.13  & 92.82 \\
 \cline{2-5}
 & {\bf Ours} & {\bf 88.42} & {\bf 92.86}& {\bf 93.13} \\
& + {\bf WCA} & 88.50 & 92.84 & 93.22\\
 & +  {\bf CL} & {\bf 88.57} & {\bf 92.94} & {\bf 93.20} \\ 
 \hline
 \hline
 \multirow{6}{*}{QNLI} & Random & 76.33 & 83.61 & 84.63\\
 & Entropy & 77.95 & 83.83 & 84.75 \\
 & BADGE & 77.74& 84.90 & 84.32 \\
 & CAL & {\bf 78.53} & {\bf 85.14} & {\bf 84.99} \\
 \cline{2-5}
 & {\bf Ours} & 78.44& 84.93 & 84.87 \\
& + {\bf WCA} & 78.47 & 85.12 & 84.91\\
 & +  {\bf CL} & {\bf 78.92} & 85.06 & 84.96 \\ 
 \hline
 \hline
 \multirow{6}{*}{QQP}& Random & 77.32 & 81.73 & 84.22\\
 & Entropy & 78.47 & 81.92 & 86.03 \\
 & BADGE & 78.02 & 81.63 & 84.06 \\
 & CAL & 78.23  & {\bf82.52}& 84.25 \\
 \cline{2-5}
 & {\bf Ours} & {\bf 78.97} &  82.43 & {\bf 84.77} \\
& + {\bf WCA} & 78.90 & 82.55 & 84.83\\
 & +  {\bf CL} & {\bf 79.32} & {\bf 82.91} & {\bf 84.95}\\ 
 \bottomrule
  \end{tabular}}
 \caption{Results of the in-domain test accuracies for different acquired dataset size. $+$ WCA refers to Ours $+$ select worst-case augmentation. $+$ CL refers to Ours $+$ curriculum learning. We
provide error bars in Table~\ref{tab:appendix_in_domain} in the Appendix. 
 }
 \vspace{-5mm}  
    \label{tab:in_domain}
\end{table}

\subsection{Out-of-Domain Classification Task Results}

We compare our proposed method with the baselines for
their performance in an out-of-domain (OD) setting and  summarize the results in Table~\ref{tab:ood_accuracy}.
We test domain generalization on three datasets with two tasks, including sentiment analysis and paraphrase detection. 
We set the annotation budget as $15\%$ of $\cD_{pool}$ for all OD experiments. For OD in SST-2 and IMDB, {\ours}  yields better results than all baselines with a clear margin ($1.7\%$ and $2.0\%$, respectively). 
With curriculum learning, the results are continually improved. The performance
gains on out-of-domain are often greater than the gains on in-domain, implying that {\ours} can significantly help the model to generalize across domains. 
On QQP, {\ours} achieves comparable results as CAL without curriculum learning while the performance can be further improved by adding curriculum learning.

\begin{table}[h]
 \footnotesize
\centering
\resizebox{0.8\columnwidth}{!}{
 \begin{tabular}{l|c|c|c}
 \toprule
ID & SST-2  & IMDB & QQP \\ 
OD & IMDB  & SST-2  & TwitterPPDB\\
 \hline
Random  & 76.31 & 82.01 & 85.57\\
Entropy & 75.88 & 85.32 & 85.18\\
BADGE  & 75.23 & 85.11 & 85.39\\
CAL & 78.88& 84.92 & {\bf 86.14}\\
\hline
{\bf {Ours}} & {\bf80.54} & {\bf 86.97}  &86.03  \\
+ {\bf WCA} & 80.72  & 86.99 & 86.07\\
+ {\bf CL} & {\bf 80.91} & {\bf 87.07} & {\bf 86.18}\\
 \bottomrule
  \end{tabular}}
  \vspace{-3mm}
 \caption{
 Results of out-of-domain (OD) generalization. We report
the out-of-domain accuracy on the target domain. ID refers to in-domain dataset. OD refers to out-of-domain dataset.}
\vspace{-6mm}  
    \label{tab:ood_accuracy}
\end{table}

\subsection{Prompt-Based Few-Shot Learning Results}\label{sec:prompt_based_few_shot}
We present the prompt-based few-shot learning results with GPT-2 in Table \ref{tab:few_shot}, in which we follow the setting (4-shot, 8-shot, and 12-shot) in \citet{zhao2021calibrate}.
Few-shot learners suffer from the quality of labeled data \citep{sohn2020fixmatch}, and previous acquisition functions usually fail to boost the performance from labeling random sampled data.
In Table \ref{tab:few_shot}, we observe that uncertain prompts performs similar to random selected prompts. A potential reason is that an under-trained model treats all examples as uncertainty examples and hard to distinguish the informativeness.
However, our proposed acquisition demonstrates the strong capability in modeling the local sensitivity and learning from easy to hard. 
It comes to the best performance in most of the settings. 
These findings show the potential of using our acquisition to improve prompt-based few-shot learning and make a good in-context examples for GPT-2 model.

\begin{table}[h]
 \footnotesize
 \vspace{-2mm}
\centering
\resizebox{0.8\columnwidth}{!}{
 \begin{tabular}{l|l|c|c|c}
 \toprule
  && \multicolumn{1}{c|}{4-shot} &\multicolumn{1}{c|}{8-shot} &\multicolumn{1}{c}{12-shot}\\ 
   \hline
\multirow{4}{*}{SST-2} & Random & 64.9 & 54.5 & 56.3\\
 & Calibrated & 73.8 & 64.6 & 73.0 \\
  & Uncertainty & 59.7 & 64.5 & 66.8  \\
\cline{2-5}
 & {\bf Ours} & {\bf 75.3} & {\bf  77.8} & {\bf 79.7} \\
 \hline
 \hline
 \multirow{4}{*}{TREC} & Random & 23.1 & 32.7 & 37.5\\
 & Calibrated & 44.2 & 44.1 & 44.4 \\
  & Uncertainty & 34.8 & 52.2 & 54.1  \\ 
 \cline{2-5} 
 & {\bf Ours} & {\bf 46.4}& {\bf 58.7} & {\bf 59.8} \\
 \hline
 \hline
 \multirow{4}{*}{RTE} &Random & 53.2 & 54.9 & 56.0\\
 &Calibrated & 57.5 & 57.7 & 58.2 \\
  &Uncertainty & 57.0 & 57.3 & 57.8\\
 \cline{2-5}
 & {\bf Ours} & {\bf 57.9} & {\bf 58.4} & {\bf 59.7} \\
 \bottomrule
  \end{tabular}}
  \vspace{-3mm}
 \caption{Results across different
strategies of acquiring training examples (the prompt format is fixed). The language model here is GPT-2 (1.5B). 
 }
 \vspace{-6mm}  
    \label{tab:few_shot}
\end{table}

\section{Analysis}\label{sec:analysis_section}
\paragraph{Can we use our proposed acquisition in the imbalance setting?}

Extreme label imbalance is an important challenge in many non-pairwise NLP tasks \cite{sun2009strategies,zhang2017position, mussmann2020importance}. We set up the imbalance setting by sampling a subset with class-imbalance sample rate. For binary classification, 
we set the positive-class data sample rate as $1.0$ and negative-class data sample rate as $0.1$. 
As our acquisition focuses on local sensitivity and informativeness, 
it tends to select examples close to the decision boundary.
Once too many positive examples and few negative examples are labeled, the local perturbation around negative samples are easy to be positive, and thus
{\ours} selects examples that are close to the negative examples. 
We conduct the experiments on SST-2, IMDB, and AG News with annotation budget as $1\%$. In Table \ref{tab:analysis_imbalance}, Ours\footnote{Ours in the Section \ref{sec:analysis_section} refers to ours + curriculum learning.} indicates strong improvements. 
This further proves that our selection method can generalize better.

\begin{table}[h]
 \footnotesize
 \vspace{-2.0mm}
\centering
\resizebox{0.65\columnwidth}{!}{
 \begin{tabular}{l|c|c|c}
 \toprule
  & {SST-2} & {IMDB} &{AG News}\\ 
 \hline
Random & 79.45   & 62.33  & 82.95\\  
Entropy  & 81.71   & 65.69 & 82.79\\ 
CAL & 83.23  & 72.75 & 83.27\\
{\bf Ours} &  {\bf 85.48} & {\bf 74.48} & {\bf 84.11}\\
 \bottomrule
  \end{tabular}}
  \vspace{-3mm}
 \caption{Main results of different active learning strategies on the imbalanced SST-2, IMDB, and AG News. }
 \vspace{-6mm}  
    \label{tab:analysis_imbalance}
\end{table}

\paragraph{Would different augmentations make meaningful difference?}
We test if our results are sensitive to the choice of augmentation: TF-IDF and backtranslation.
For TF-IDF, we compare the random sample augmentation and worst-case augmentation (WCA).
TF-IDF and Backtranslation generate diverse paraphrases while preserving the semantics
of the original sentences. Select-worst case augments the inputs by incorporating the approximate adversarial perturbations. 
Table \ref{tab:analysis_augmentation} indicates our method is insensitive to different augmentations.
We also observe that WCA achieves the highest gains on two datasets. This confirms our discussion in Section \ref{sec:acquisition_with_local} that  select-worst case is capable of imposing  local sensitivity.

\begin{table}[t]
 \footnotesize
\centering
\resizebox{0.7\columnwidth}{!}{
 \begin{tabular}{l|c|c|c}
 \toprule
  & {SST-2} & {IMDB} &{AG News}\\ 
 \hline
Backtranslation & 86.01  & 75.12 & 88.39 \\
TF-IDF & 85.97  & 75.23 & 88.42  \\  
+ WCA  & 86.37 &  75.17 & 88.50\\ 
 \bottomrule
  \end{tabular}}
  \vspace{-3mm}
 \caption{Acquisition performance for different augmentations. We report results of our acquisition with different augmentations to get the local copies of the samples. }
 \vspace{-6mm}  
    \label{tab:analysis_augmentation}
\end{table}

\paragraph{What is the influence of the choice of divergence?}
We select different divergences in the statistical distance family and 
study their abilities in encoding different information. 
Corresponding to Section \ref{sec:acquisition_with_local}, we present the results in Table \ref{tab:analysis_divergence}. We experiment on the  KL divergence, JSD, and $\alpha$-divergence \cite{minka2005divergence} with the $\alpha$ value  set as $-0.5$ or $0.5$.
We notice that for our case the difference between different divergences is small. A possible reason is that the number of class categories is small and therefore the choice of divergence does not have a large influence.





\paragraph{Can we use the proposed acquisition with extremely few labeled data?}
We have presented the results under very limited annotation budgets in Table \ref{tab:in_domain}.
 We set the annotation budget as $0.8\%$ and $0.4\%$. The key observation is that the degradation of performance in the other acquisition functions are dramatic. For example, in IMDB, the uncertainty sampling (Entropy) shows the obvious performance drop. It suffers from the sampling bias problem because of the frequent variation of the decision boundary in the early phase of
training with very few labeled data available, which results in ineffective sampling. Even under this extreme case, our acquisition still aims to select the most informative examples for the model.
This further verifies our empirical results in Section \ref{sec:prompt_based_few_shot} on prompt-based few-shot learning where only a very few in-context prompts are provided.

\section{Related Work}

\paragraph{Active Learning} 
Active Learning has been widely used in many applications in NLP \cite{lowell2018practical,dor2020active,ru2020active}. The uncertainty-based methods \cite{fletcher2008resolution} 
\begin{table}[h]
 \footnotesize
\centering
\resizebox{0.7\columnwidth}{!}{
 \begin{tabular}{l|c|c|c}
 \toprule
  & {SST-2} & {IMDB} &{AG News}\\ 
 \hline
KL & 86.37  & 77.57 & 88.57  \\  
JSD  & 86.25  & 77.38 & 88.41\\ 
$\alpha =-0.5$ & 86.31  & 77.42 & 88.43 \\
$\alpha =0.5$ & 86.39  & 77.53 & 88.61 \\
 \bottomrule
  \end{tabular}}
  \vspace{-3mm}
 \caption{Ablation study on different choices of divergences. We report KL, JSD, and $
\alpha$-divergence, and set $\alpha = \pm 0.5$ respectively.}
\vspace{-3mm}  
    \label{tab:analysis_divergence}
\end{table}
\begin{table}[h]
 \footnotesize
 \vspace{-0.0mm}
\centering
\resizebox{0.7\columnwidth}{!}{
 \begin{tabular}{@{}lccccc@{}}
 \toprule
 \small  & \multicolumn{2}{c}{\small SST-2} & 
& \multicolumn{2}{c}{\small IMDB } \\
\cmidrule{2-3} \cmidrule{5-6}
  Dataset size& {0.4\%} & {0.8\%} && {0.4\%} & {0.8\%} \\ 
 \hline
Random & 64.64   & 61.08 && 60.84 & 73.86\\  
Entropy  & 67.88   & 63.94 && 58.96  & 71.32\\ 
CAL & 73.81  & 65.72 && 61.65 & 74.15\\
{\bf Ours} &  {\bf 76.45} & {\bf 69.46} && {\bf 64.54}& {\bf 75.88} \\
 \bottomrule
  \end{tabular}}
  \vspace{-3mm}
 \caption{Results on the SST-2 and IMDB datasets under limited annotation budget (0.4\%, 0.8\%). 
}
\vspace{-6.5mm}  
    \label{tab:analysis_extreme_limited}
\end{table}
have become the most common strategy. Instead of only considering uncertainty, diversity sampling 
has also become an alternative direction. Recent works \cite{geifman2017deep,sener2017active,Ash2020Deep, yuan2020cold} focus on  different parts of  diversity. 
Most recent works \citep[$e.g.$][]{zhang2021cartography,margatina2021active} have been more on exploiting the model behavior  and each individual instance. Our work focuses more on the local sensitivity and informativeness of data, leading to better performance under various limited annotation settings.

\paragraph{Annotation Budgeting}
Annotation budgeting with learning has long been studied~\cite{Turney2002TypesOC}. \citet{Sheng2008GetAL} study the tradeoff between collecting multiple labels per example versus annotating more examples. 
On the other hand, different labeling strategies such as providing fine-grained rationales~\cite{Dua2020BenefitsOI}, active learning~\cite{kirsch2019batchbald}, and the training dynamics approach ~\cite{Swayamdipta2020DatasetCM} are studied. 
Except standard classification, class-imbalance \cite{Mussmann2020OnTI} or noisy label cases \cite{fan2021contextual,chen2021learning} have also been explored. 
We utilize active learning to explore the labeling strategies and aim to select the most informative data for annotations.

\section{Conclusion}\label{sec:conclusion}

Our work demonstrates the benefits of introducing local sensitivity and learning from easy to hard into the acquisition strategy. The proposed acquisition function 
shows noticeable gains in performance across classification tasks and prompt-based few-shot learning. In this work, we conduct the detailed study with  the proposed acquisition strategy in different settings, including imbalanced  and extremely limited labels.
We also verify the impact of  different choice of designs such as the choice of divergence and augmentations. 
To summarize, the proposed {\ours} is effective and general, with the potential to be incorporated into existing models for various NLP tasks.

\section*{Acknowledgements}
S. Zhang and M. Zhou acknowledge the support of NSF IIS-1812699 and Texas Advanced Computing Center.

\bibliography{naacl2021}
\bibliographystyle{acl_natbib}
\clearpage

\clearpage
\appendix
\section{Experimental details}\label{sec:app_exp}
\subsection{Full Results and Examples}\label{sec:appendix_fullresults}
We report the full results of  out-of-domain and in-domain  tasks in Tables \ref{tab:appendix_ood_accuracy} and \ref{tab:appendix_in_domain}, respectively. The full results of prompt-based few-shot learning are shown in Table \ref{tab:appendix_few_shot} and Table \ref{tab:prompt_examples} shows prompt examples of each task.
\begin{table}[h]
 \footnotesize
\centering
 \begin{tabular}{l|c|c|c}
 \toprule
ID & SST-2  & IMDB & QQP \\ 
OD & IMDB  & SST-2  & TwitterPPDB\\
 \hline
Random  & 76.31$\pm$0.66 & 82.01$\pm$3.45 & 85.57$\pm$0.42\\
Entropy & 75.88$\pm$1.82 & 85.32$\pm$2.36 & 85.18$\pm$1.79\\
BADGE  & 75.23$\pm$0.87 & 85.11$\pm$2.92 & 85.39$\pm$3.44\\
CAL & 78.88$\pm$1.27& 84.92$\pm$2.30 & 86.14$\pm$0.31\\
\hline
{\bf {Ours}} & 80.24$\pm$0.91 & 86.07$\pm$2.45  &86.03$\pm$0.40  \\
+ {\bf WCA} & 80.42$\pm$0.85  & 86.19$\pm$2.37 & 86.07$\pm$0.36\\
+ {\bf CL} & 80.51$\pm$0.67 & 86.24$\pm$1.98 & 86.18$\pm$0.29\\
 \bottomrule
  \end{tabular}
 \caption{
 Results of out-of-domain (OD) generalization. We report
the out-of-domain accuracy on the target domain. ID refers to in-domain dataset. OD refers to out-of-domain dataset.}
    \label{tab:appendix_ood_accuracy}
\end{table}

\begin{table}[h]
 \footnotesize
\centering
\resizebox{1.0\columnwidth}{!}{
 \begin{tabular}{l|l|c|c|c}
 \toprule
  && \multicolumn{1}{c|}{4-shot} &\multicolumn{1}{c|}{8-shot} &\multicolumn{1}{c}{12-shot}\\ 
   \hline
 \multirow{4}{*}{SST-2} & Random & 64.9$\pm$8.4 & 54.5$\pm$4.6 & 56.3$\pm$2.3 \\
 & Calibrated & 73.8$\pm$10.9 & 64.6$\pm$8.8 & 73.0 $\pm$5.3\\
  & Uncertainty & 59.7$\pm$7.3 & 64.5$\pm$5.9 & 66.8$\pm$4.8  \\
\cline{2-5}
 & {\bf Ours} & 75.3$\pm$7.8 & 77.8$\pm$4.7 & 79.7$\pm$3.2 \\
 \hline
 \hline
 \multirow{4}{*}{TREC} & Random & 23.1$\pm$5.9 & 32.7$\pm$7.5 & 37.5$\pm$7.8\\
 & Calibrated & 44.2$\pm$2.2 & 44.1$\pm$3.6 & 44.4$\pm$4.0 \\
  & Uncertainty & 34.8$\pm$3.4 & 52.2$\pm$4.1 & 54.1$\pm$5.2  \\
 \cline{2-5}
 & {\bf Ours} & 46.4$\pm$2.8 &  58.7$\pm$3.6 & 59.8$\pm$4.3 \\
 \hline
 \hline
 \multirow{4}{*}{RTE} &Random & 53.2$\pm$6.0 & 54.9$\pm$3.0 & 56.0$\pm$2.2\\
 &Calibrated & 57.5$\pm$1.8 & 57.7$\pm$1.3 & 58.2$\pm$1.1 \\
  &Uncertainty & 57.0$\pm$1.5 & 57.3$\pm$1.4 & 57.8$\pm$1.1\\
 \cline{2-5}
 & {\bf Ours} & 57.9$\pm$2.3 &  58.4$\pm$1.6 & 59.7$\pm$1.2 \\
 \bottomrule
  \end{tabular}}
 \caption{Full results across different
choices of the training examples (the prompt format is fixed). The language model at here is GPT-2XL (1.5B). 
 }
    \label{tab:appendix_few_shot}
\end{table}

\begin{table}[h]
 \footnotesize
\centering
\resizebox{1.0\columnwidth}{!}{
 \begin{tabular}{l|l|c|c|c}
 \toprule
  &Acquired dataset size:& \multicolumn{1}{c|}{1\%} &\multicolumn{1}{c|}{5\%} &\multicolumn{1}{c}{10\%}\\ 
   \hline
 \multirow{6}{*}{SST-2} & Random & 84.11$\pm$0.45 & 86.53$\pm$0.61 & 88.05$\pm$0.73\\
 & Entropy & 84.53$\pm$0.81 & 87.82$\pm$0.73 & 89.45$\pm$0.92 \\
  & BADGE & 84.32$\pm$0.64 & 87.11$\pm$0.82 & 88.72$\pm$0.44 \\
 & CAL & 84.95$\pm$0.56 & 87.34$\pm$0.61 & 89.16$\pm$0.67 \\
\cline{2-5}
 & {\bf Ours} & 85.97$\pm$0.53 & 88.61$\pm$0.48 & 90.05$\pm$0.61 \\
& + {\bf WCA} & 86.12$\pm$0.47 & 88.56$\pm$0.55 & 90.14$\pm$0.57\\
 & + {\bf CL} & 86.37$\pm$0.43 & 88.79$\pm$0.46 & 90.18$\pm$0.48\\ 
  
 \hline
 \hline
 \multirow{6}{*}{IMDB}& Random & 65.96$\pm$0.66 & 84.22$\pm$0.52 & 86.25$\pm$0.54\\
 & Entropy & 68.32$\pm$0.53 & 84.51$\pm$0.48 & 87.29$\pm$0.51 \\
 & BADGE & 67.80$\pm$0.44 & 84.46$\pm$0.50 & 87.17$\pm$0.41 \\
 & CAL & 73.55$\pm$0.56 & 84.72$\pm$0.48 & 87.27$\pm$0.50 \\
 \cline{2-5}
 & {\bf Ours:} & 75.23$\pm$0.43 & 85.82$\pm$0.35 & 87.91$\pm$0.53\\
& + {\bf WCA} & 75.17$\pm$0.58 & 85.79$\pm$0.67 & 87.83$\pm$0.71\\
 & +  {\bf CL} & 77.57$\pm$0.64 & 86.02$\pm$0.62 & 88.43$\pm$0.57 \\ 
 \hline
 \hline
 \multirow{6}{*}{AG News} &Random & 85.43$\pm$0.53 & 90.05$\pm$0.51 & 91.93$\pm$0.60\\
 &Entropy & 86.48$\pm$0.46 & 92.21$\pm$0.41 & 92.65$\pm$0.39 \\
 &BADGE & 86.81$\pm$0.48 & 90.72$\pm$0.51 & 92.41$\pm$0.53 \\
 & CAL & 87.12$\pm$0.31 & 92.13$\pm$0.38 & 92.82$\pm$0.35 \\
 \cline{2-5}
 & {\bf Ours} & 88.42$\pm$0.37 & 92.86$\pm$0.40 & 93.13$\pm$0.39 \\
& + {\bf WCA} & 88.50$\pm$0.35 & 92.84$\pm$0.37 & 93.22$\pm$0.42\\
 & +  {\bf CL} & 88.57$\pm$0.30 & 92.94$\pm$0.32 & 93.20$\pm$0.35 \\ 
 \hline
 \hline
 \multirow{6}{*}{QNLI} & Random & 76.33$\pm$0.54 & 83.61$\pm$0.57 & 84.63$\pm$0.62\\
 & Entropy & 77.95$\pm$0.50 & 83.83$\pm$0.61 & 84.75$\pm$0.55 \\
 & BADGE & 77.74$\pm$0.53 & 84.90$\pm$0.48 & 84.32$\pm$0.46 \\
 & CAL & 78.53$\pm$0.49 & 85.14$\pm$0.45 & 84.99$\pm$0.53 \\
 \cline{2-5}
 & {\bf Ours} & 78.44$\pm$0.41 & 84.93$\pm$0.32 & 84.87$\pm$0.39 \\
& + {\bf WCA} & 78.47$\pm$0.43 & 85.12$\pm$0.37 & 84.91$\pm$0.38\\
 & +  {\bf CL} & 78.92$\pm$s0.40 & 85.06$\pm$0.36 & 84.96$\pm$0.33 \\ 
 \hline
 \hline
 \multirow{6}{*}{QQP}& Random & 77.32$\pm$0.66 & 81.73$\pm$0.72 & 84.22$\pm$0.75\\
 & Entropy & 78.47$\pm$0.57 & 81.92$\pm$0.64 & 86.03$\pm$0.49 \\
 & BADGE & 78.02$\pm$0.49 & 81.63$\pm$0.55 & 84.06$\pm$0.60 \\
 & CAL & 78.23$\pm$0.52 & 82.52$\pm$0.57 & 84.25$\pm$0.48 \\
 \cline{2-5}
 & {\bf Ours} & 78.97$\pm$0.46 &  82.43$\pm$0.44 & 84.77$\pm$0.52 \\
& + {\bf WCA} & 78.90$\pm$0.50 & 82.55$\pm$0.48 & 84.83$\pm$0.48\\
 & +  {\bf CL} & 79.32$\pm$0.53 & 82.91$\pm$0.51 & 84.95$\pm$0.58\\ 

 \bottomrule
  \end{tabular}}
 \caption{Full results of the in-domain test accuracies for different acquired dataset size. $+$ WCA refers to Ours $+$ select worst-case augmentation. $+$ CL refers to Ours $+$ curriculum learning.  }
    \label{tab:appendix_in_domain}
\end{table}

\begin{table*}[t]

\centering
\resizebox{2\columnwidth}{!}{
 \begin{tabular}{lll}
 \toprule
  Task & Prompt & Label Names  \\
 \hline
 SST-2  &  Review: At times, the movie looks genuinely pretty. & Positive, Negative\\
 & Sentiment: Positive \\[+2.0ex]
 & Review: The movie is amateurish, but it's a minor treat. \\
 & Sentiment:\\
 \hline
TREC & Question: Where can I find information on becoming a journalist? & Number, Location, Person, Description,  \\
& Answer Type: Location & Entity, Abbreviation\\ [+2.0ex]
& Question: What is the temperature today? \\
& Answer Type:

\\ \hline
RTE &  The motor industry accounts for as much as 40 percent of the 450,000 installed industrial robots & True, False \\
&worldwide but their use is changing and applications are expanding. \\
& Question: The most common use for robots is the manufacture of automobiles. True or False? \\
& Answer: True \\ [+2.0ex]

 & Arroyo was the favorite of investors because of her experience as a trained economist  \\
 & and government manager.\\
 & Question: Arroyo has experience as an economist and as a government manager. True or False? \\
 & Answer: \\
 \bottomrule
  \end{tabular}}
 \caption{The different prompts we use for SST-2, TREC, and RTE. One training example per task is presented. The language model is used to predicted the label probability as shown in the right column. }
    \label{tab:prompt_examples}
\end{table*}

\subsection{Classification Task Hyperparameters and Experimental Settings}
Our implementation is based on the BERT-base \cite{devlin2018bert} from \textit{HuggingFace Transformers}~\citep{wolf2020transformers}.
We optimize the KL divergence as the objective with the Adam optimizer \citep{kingma2014adam} and batch size is set to 16 for all experiments.
The curriculum learning is trained for $200$ iterations.
The learning rate is $2 \times 10^{-5}$. The $\alpha$ in Eqn \eqref{eq:ssl} is set as $0.01$ for all experiments.  With
longer input texts such as IMDB, we use $256$ as the maximum sequence length. For others, we use  $128$. Following \citet{Ash2020Deep} and \citet{margatina2021active}, for the initial training set $\cD_{label}$, we begin the active learning loop by uniformly random sampling from $\cD_{pool}$. For all experiments in the Section \ref{sec:analysis_section}, we set the annotation budget as $1\%$ and use Ours (ours + curriculum learning) as the default methods. 

TF-IDF based data augmentation \cite{Xie2020UnsupervisedDA} aims to generate
both diverse and valid examples. It is designed to retain keywords and replace uninformative words with other uninformative words. BERT is used as the word tokenizer. 
We set IDF$(w)$ is the IDF score for word $w$ computed on the whole corpus, and TF$(w)$ is the TF score for word $w$ in a sentence. Then, we compute the TF-IDF score as TFIDF$(w) =$ TF$(w)$IDF$(w)$. Suppose the maximum TF-IDF score in a sentence $\xv$ is C = max$_i$ TFIDF$(x_i)$.
We set the probability to min($p$(C - TFIDF$(x_i)$)/Z, 1), where $p$ is a hyperparameter that controls the
magnitude of the augmentation and we set $p=$0.3. Z is the average score over the inputs sentence. For backtranslation, we use a pre-trained EN-DE\footnote{\url{https://dl.fbaipublicfiles.com/
fairseq/models/wmt19.en-de.joined-dict.
single_model.tar.gz}} and DE-EN\footnote{\url{https://dl.fbaipublicfiles.com/
fairseq/models/wmt19.de-en.joined-dict.
single_model.tar.gz}} translation
models \cite{ng2019facebook} to perform backtranslation on each sentence.




\subsection{Prompt-based Few-Shot Learning Hyperparameters and Experimental Settings}
We use the 1.5B parameters GPT-2 \cite{radford2019language}, with a Transformer \cite{vaswani2017attention} based architecture. The model largely follows the details
of the OpenAI GPT model \cite{radford2018improving} with a few modifications. 
Layer normalization \cite{ba2016layer, fan2020bayesian, zhang2021bayesian}
is moved to the input of each sub-block and an
additional layer normalization is added after the final self-attention block. 
Following the settings in \citet{zhao2021calibrate}, the maximum input length is 2048 tokens or 1500 words. In Table \ref{tab:prompt_examples}, we show the default prompt format for SST-2, TREC, and RTE. 
For datasets, Stanford Sentiment Treebank (SST-2) \cite{socher2013recursive} is one of benchmarks in General Language Understanding Evaluation (GLUE) \cite{wang2018glue}. With fully labeled parse tress, This corpus allows a complete analysis of the compositional effects of sentiment in language. TREC \cite{voorhees2000building} is a 6-way question classification. The target is to classify the questions based on whether their answer type is a Number, Location, Person, Description, Entity, or Abbreviation. Similarly, RTE (Recognizing Textual Entailment) \cite{dagan2005pascal} is also a benchmark dataset from GLUE. It is a binary classification task to determine if
a given premise entails a given hypothesis.

\subsection{Sampling Efficiency and Running Time}\label{sec:appendix_runningtime}

\begin{table}[h]
 \footnotesize
\centering
 \begin{tabular}{l|c|c|c|c}
 \toprule
  & {SST-2} & {IMDB} &{AG News} & {AVG.}\\ 
 \hline
Random & 0 & 0  & 0 & 0\\  
Entropy  & 173  & 107 & 402 & 227\\
BADGE  & 25640 & 3816 & 1961 & 10303\\
CAL & 708  & 273 & 1284 & 755 \\
 \hline
Ours & 513 & 228 & 881 & 541 \\
+ {WCA} & 611 & 275 & 1023 & 636\\
 \bottomrule
  \end{tabular}
 \caption{Running time (seconds) per sampling iteration (inference and selection) during AL acquisition for each datasets. AVG. refers the average acquisition time for all three datasets. For each acquisition, we report the running time under three adversarial attacks respectively}
\label{tab:appendix_runningtime}
\end{table}


We mask $m$ as the number of labeled data in $\cD_{label}$, $n$ the number of unlabeled data in $\cD_{pool}$, $C$ the
number of classes in the downstream classification
task, $d$ the dimension of embeddings, $l$ the maximum sequence length, and $s_\textit{acq}$ the acquisition size. 
We set these values following \citet{yuan2020cold} and \citet{ margatina2021active}.
In Table~\ref{tab:appendix_runningtime}, running time in seconds are summarized per sampling iteration (inference and selection) during AL acquisition for each dataset. Experiments in this part
are performed on a Tesla V100 GPU. We keep $s_\textit{acq} = 100, d = 768, t = 10$, and $l = 128$. 
For IMDB, we change the maximum sequence length to $256$. 
As demonstrated in Table~\ref{tab:appendix_runningtime}, BADGE requires a significantly amount of running time, since it has to cluster high-dimensional vectors and is a very computationally-heavy method.  
CAL also requires relative long running time as it needs to find the contrastive examples by finding nearest neighbors and computing contrastive score for unlabeled candidates. Our method achieves the second best efficiency. Even with the select worst-case augmentation, our acquisition function is still computationally productive as the augmentation and ranking candidates can be well deployed in the current computational machines.
Entropy is overall the most efficient method as it only requires to rank the list of uncertainty scores, while it tends to have weaker performance.

\end{document}